\documentclass[conference]{IEEEtran}
\IEEEoverridecommandlockouts
\usepackage{cite}
\usepackage{amsmath,amssymb,amsfonts}
\usepackage{algorithmic}
\usepackage{graphicx}
\usepackage{textcomp}
\usepackage{tabularx}
\usepackage{booktabs}
\usepackage{xcolor}
\def\BibTeX{{\rm B\kern-.05em{\sc i\kern-.025em b}\kern-.08em
    T\kern-.1667em\lower.7ex\hbox{E}\kern-.125emX}}
\begin{document}

\title{Evaluating AI Generated Summaries for Cancer Patients}

\author{
\IEEEauthorblockN{
Muhammad Aurangzeb Ahmad,
Kim Shyu,
Leon Oliver,
Fergus Sleight,
Paul Landau
}

\IEEEauthorblockA{
Careology Inc.\\
\{muhammad, kim, leon, fergus, paul\}@careology.health
}
}

\maketitle

\begin{abstract}
Large language models (LLMs) are increasingly being integrated into digital health platforms to generate summaries of complex medical data. Although these models can improve patient engagement and communication, these systems also raise concerns about accuracy, faithfulness, and safety in clinical contexts. In this study, we evaluate AI-generated summaries within a cancer patient care application using a dual assessment framework. Human domain experts, including oncology clinicians and patient-facing care staff, provided ground-truth evaluations of summary quality along dimensions of accuracy, clinical relevance, and readability. In parallel, we employed LLMs serving as evaluators (\textit{LLM-as-a-judge}). Some limitations were identified in the generated summaries e.g., occasional omissions and minor inaccuracies. These were systematically analyzed and used to iteratively improve prompt design, grounding, and safety guardrails.
\end{abstract}

\begin{IEEEkeywords}
Large language models (LLMs), healthcare informatics, clinical summarization, oncology care platforms, LLM-as-a-judge, human-centered evaluation, AI safety in healthcare
\end{IEEEkeywords}

\section{Introduction}
Healthcare systems generate large volumes of clinical documentation across encounters, care settings, and time. Clinicians must routinely synthesize this information to support decision-making, care coordination, and communication with patients and caregivers. However, electronic health records (EHRs) often fragment clinically relevant information across progress notes, discharge summaries, orders, and ancillary documentation, increasing cognitive burden and contributing to information overload \cite{gal2021quantifying, asgari2024impact}. Recent advances in large language models (LLMs) have enabled automated generation of concise, narrative summaries from heterogeneous clinical data \cite{topol2019high, singhal2023large}. AI-generated patient summaries are increasingly explored as clinician-facing tools intended to support chart review, care transitions, handoffs, and preparation for patient communication. In this context, summaries are not delivered directly to patients, but rather serve as synthesized representations of patient status that clinicians may review, edit, or incorporate into downstream documentation and discussions.

Despite their potential to improve efficiency and situational awareness, LLM-generated summaries introduce distinct clinical risks. Errors or omissions may propagate through clinical workflows, influencing clinician understanding, documentation, or communication with patients. Unlike traditional extractive summaries, generative models may introduce unsupported statements, misstate numerical values, or subtly reframe clinical priorities \cite{maynez2020faithfulness}. Because these summaries may be reused across encounters or inform subsequent documentation, even small inaccuracies can accumulate or persist over time which may result in patient harm. Thus, there is a need to evaluate patient generated summaries in a real world settings.

Most existing work on LLMs in healthcare has focused on clinician-facing documentation assistance, medical question answering, or decision support. AI-generated patient summaries occupy a unique position within clinical workflows since they are neither raw clinical notes nor final patient-facing documents. Instead, they function as intermediate artifacts that shape clinician interpretation and downstream communication. This intermediary role implies that the summaries must have consistency with source data, alignment with clinician intent, and there should be transparent regarding uncertainty and model limitations \cite{sendak2020path}. Evaluating generative models for this use case remains challenging. Conventional natural language generation metrics do not capture clinically relevant failure modes such as omission of key diagnoses, inconsistency across time, or misrepresentation of care plans \cite{novikova2017we}. Additionally, limited guidance exists on how to monitor these systems after deployment, particularly with respect to drift in model behavior, changes in underlying data distributions, and differential performance across patient subgroups \cite{khattak2024mlhops}.

In this work, we examine AI-generated patient summaries as a clinician-facing, safety-critical application of generative AI. We focus on the evaluation, governance, and monitoring practices required for responsible deployment of AI generated summaries within clinical environments. We employ a framework that focuses on clinically meaningful quality dimensions like 
Coherency, Fluency, Consistency, Relevance and Clinical use to asses the quality of such summaries. We also explore scenarios where the LLM-as-a-judge may be less accurate than the LLM being used to generate the summaries. The quality of the LLM generated summaries is assessed by the clinical staff at Guy's and St Thomas' Hospital. By assessing AI generated summaries using human-in-the-loop as well as LLM-as-a-judge in a production environment, we demonstrate how insights from such evaluations can be used to improve deployment and usage of AI generated summaries in a clinical context.

\section{Related Work}
Early work on clinical summarization focused on template-based approaches applied to structured or semi-structured electronic health record (EHR) data, including discharge summaries, progress notes, and radiology reports \cite{luhn1958automatic, mishra2014text}. These systems emphasized factual accuracy and traceability but were often limited in flexibility and personalization. More recent approaches leverage neural  summarization models trained on large clinical corpora, demonstrating improved fluency and coherence \cite{kanwal2022attention, zhang2020pegasus}. However, multiple evaluations have highlighted trade-offs between linguistic quality and faithfulness, including omission of clinically salient details and hallucinated content \cite{maynez2020faithfulness}. With the emergence of large language models, several studies have explored zero-shot or few-shot summarization of clinical notes \cite{singhal2023large}. While these models exhibit strong surface-level performance, empirical evaluations show risks related to numerical errors, unsupported assertions, and variability across prompts \cite{kim2025medical}. These findings underscore the need for domain-specific constraints and evaluation frameworks when deploying LLM-based summarization in clinical contexts.

A parallel body of work examines the transformation of clinician-authored documentation into patient-accessible language. Research on plain-language discharge instructions, after-visit summaries, and shared decision-making tools demonstrates that readability and relevance significantly impact patient comprehension, adherence, and satisfaction \cite{weiss2003health, burns2022readability}. Recent LLM-based systems show promise in generating more natural and empathetic explanations \cite{bedi2025testing}. Patient-facing summaries introduce distinct safety considerations, as errors may directly influence patient understanding and behavior outside clinical supervision. Beyond usage by patients, the deployment of LLMs in healthcare has expanded to applications ranging from clinical documentation assistance to decision support and patient engagement \cite{topol2019high, shandhi2022ai}. It should be noted that  real-world evaluations of such summaries emphasize that model outputs must be treated as assistive rather than authoritative \cite{sendak2020path}.

AI generated summaries can lead to hallucinations, factual inaccuracy, and omissions. Human-in-the-loop designs, retrieval-augmented generation, and bounded knowledge sources have emerged as common mitigation strategies \cite{lewis2020retrieval} for such problems. It should however be noted that evaluating generative models in healthcare creates its own set of unique challenges e.g., Traditional NLP metrics such as ROUGE and BLEU correlate poorly with clinical usefulness and safety \cite{novikova2017we}. As a result, recent work emphasizes human evaluation along dimensions such as factuality, relevance, and actionability, as well as task-specific error taxonomies for hallucinations and omissions \cite{pagnoni2021understanding}. In healthcare settings, empirical evaluations of large language models have demonstrated clinically significant hallucination rates and variable performance across specialties. Studies have highlighted risks when such systems are used for patient-facing communication or clinical decision support \cite{rajkomar2018ensuring}. Studies assessing medical question answering and discharge summary generation further show that correctness and potential for harm must be explicitly evaluated by clinicians rather than inferred from automated scores \cite{wiens2019no}.

Bias and fairness in healthcare is necessary for LLM outputs have also received increasing attention. Studies document differential performance across demographic groups, clinical conditions, and language varieties \cite{suenghataiphorn2025bias}. In patient-facing applications, such disparities risk exacerbating existing health inequities, motivating bias audits, subgroup analysis, and post-deployment monitoring for drift and unintended behavior \cite{wachter2017right}. Additionally, training data derived from electronic health records and clinical documentation may encode historical inequities in access, diagnosis, and treatment, which can be amplified by generative systems if not explicitly addressed \cite{obermeyer2019dissecting}. Language models may also underperform for patients with limited English proficiency or for culturally nuanced health narratives, raising concerns about equitable communication and comprehension \cite{rajkomar2018ensuring}. Consequently, fairness-aware evaluation protocols in healthcare increasingly incorporate stratified performance reporting, counterfactual testing, and stakeholder engagement to ensure that LLM-based systems do not systematically disadvantage vulnerable populations \cite{ahmad2023creating}.

\begin{figure}[t]
    \centering
    \includegraphics[width=\linewidth]{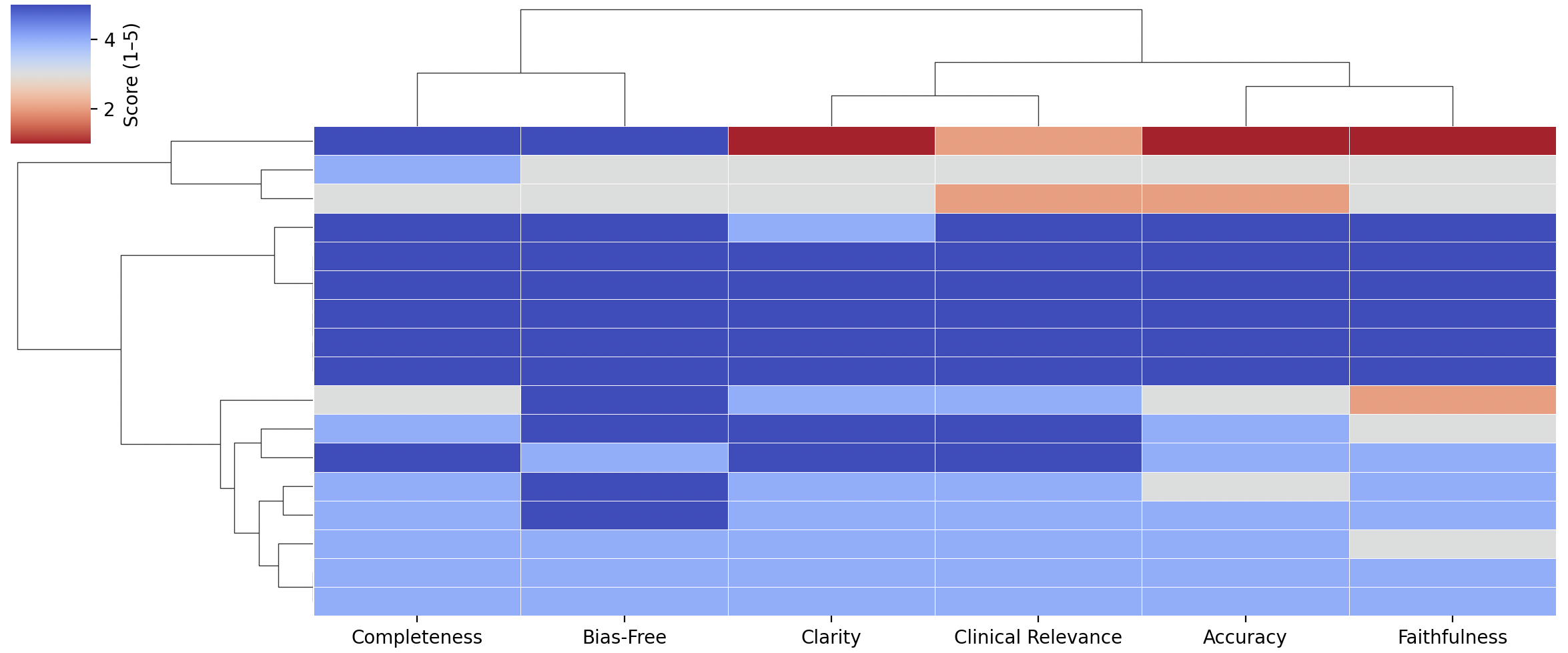}
    \caption{Hierarchical clustering of evaluation scores highlights three distinct quality phenotypes of AI-generated summaries}
    \label{fig:hier}
\end{figure}

\section{Cancer Care Monitoring App}
The setting for the AI generated summaries is a cancer care app which is part of a digital cancer care coordination and remote monitoring platform designed to support patients, caregivers, and oncology teams throughout the treatment journey. The app enables patients to report symptoms, track medications, log vital signs, and document mood or well-being in real time. Patients also have the option to add free text journal entries. In summary, the data from the app is a longitudinal record of their health outside the hospital setting. Collecting longitudinal patient-reported and physiologic data outside traditional hospital touch-points is valuable because it captures the lived experience of disease and treatment in real-world contexts that are otherwise invisible to clinicians. Many clinically meaningful events e.g., early toxicity signals, functional decline, medication non-adherence, mood deterioration, or subtle symptom escalation occur between appointments. Continuous and structured remote data collection like this reduces recall bias, enables earlier detection of deterioration, and supports proactive intervention rather than reactive care. From a systems perspective, such data also enrich predictive modeling, improve risk stratification, and allow care pathways to be optimized based on real-world trajectories rather than episodic snapshots taken during clinic visits \cite{manz2024association}. Given the volume of the data, its summarization becomes a useful overview for the clinicians.

\section{Evaluation Methodology}
Patient reported data consisting of medications, symptoms, weight, vitals (heart rate, blood pressure, temperature, oxygen saturation), mood, and free-text journals etc.  is input into an LLM (Claude Sonnet 3.7) which generates a summary based on a pre-described prompt. The summary is then validated by human domain experts as well as via automated validation (LLM-as-a-judge). We worked with Guy's and St Thomas' Hospital that is part of the NHS Trust in the United Kingdom to do the validation. A group of clinicians from the hospital volunteered to support testing. The group included a cross-functional group of domain experts: Acute Oncology Assessment Clinician, Breast Clinical Nurse Specialist, Dermatology Oncologist, Members of the Geriatric Oncology Liaison Development (GOLD) team and Oncology Physiotherapists. The experts were required to complete a review and rate AI-generated summaries for various patients. 

The experts were required to navigate to a patient profile, read the AI generated summary, and manually review the patient's data. They were then required to rate the summary across multiple criteria. Experts were asked to rate the results of each individual test case on a scale of 1-5, where 1 is extremely low and 5 is extremely high for the following criteria:

\begin{figure}[t]
    \centering
    \includegraphics[width=\linewidth]{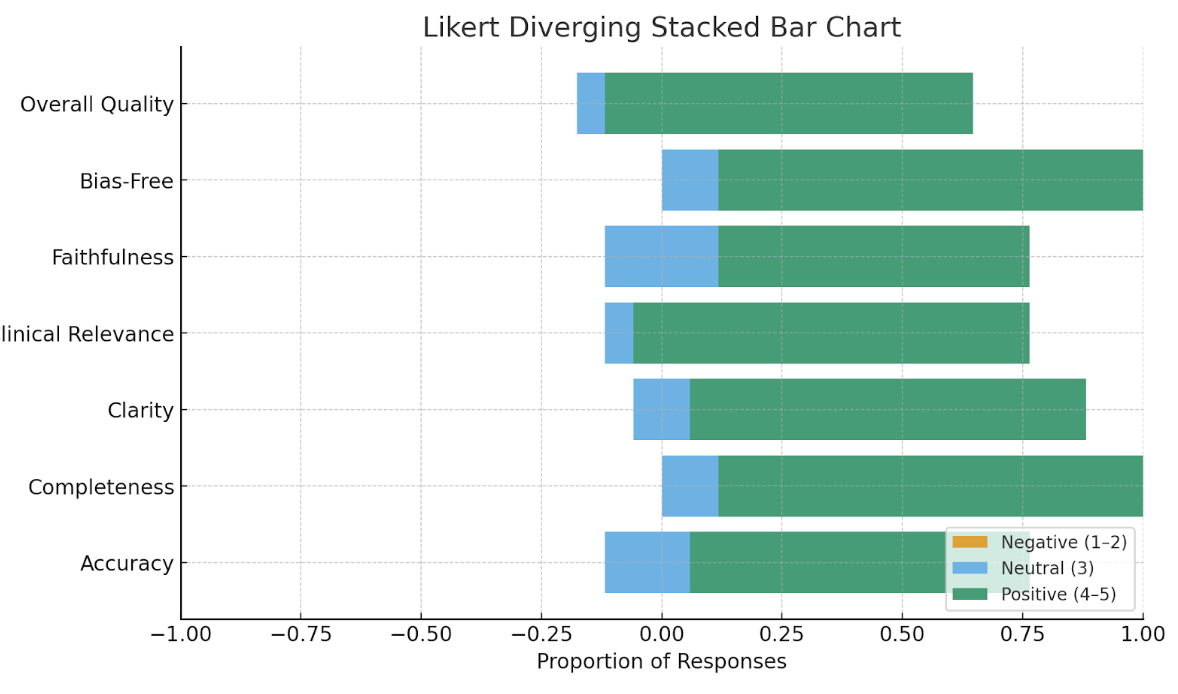}
    \caption{Diverging stacked bar chart showing the distribution of expert ratings across evaluation dimensions for the AI-generated clinical summaries, with responses grouped into negative (1–2), neutral (3), and positive (4–5) categories.}
    \label{fig:likert_eval}
\end{figure}

\begin{itemize}
    \item \textbf{Accuracy:} The summary correctly reflects the patient’s clinical facts without errors.
    \item \textbf{Completeness:} Includes all clinically relevant details.
    \item \textbf{Clarity:} Easy to understand, unambiguous.
    \item \textbf{Clinical Relevance:} Focuses on information important for decision-making.
    \item \textbf{Faithfulness:} No unsupported inferences beyond the data.
    \item \textbf{Bias-Free:} Avoids stigmatizing or culturally insensitive statements.
\end{itemize}
Additionally, there were also open ended questions where they could provide their feedback. 
\begin{itemize}
    \item Did the summary omit any important clinical details? 	What important clinical details did the summary omit?	
    \item Did the summary make any overgeneralizations or assumptions that were not supported by the data?	What overgeneralizations or assumptions did the summary make that were not supported by the data?	
    \item Did the summary contain any factual errors? 	What factual errors did the summary contain?	
    \item Did the summary use any ambiguous or confusing language?	What ambiguous or confusing language did the summary contain?	
    \item Did the summary include any potentially unsafe statements?	
    \item What potentially unsafe statements did the summary contain?	Overall Quality Rating (1-5)	
    \item Would you trust this summary in a real clinical setting?	Strengths of the summary	Areas needing improvement	Potential risks if used in clinical decision-making	Additional comments or suggestions for the AI system
\end{itemize}

The experts also gave an overall rating for the summary which can be thoughts of as whether the summary met essential correctness and safety expectations. In instances where the summary may not meet the criteria e.g., critical errors such as factual inaccuracy or inappropriate recommendations, experts were encouraged to document the issue with brief descriptions. After completing their assigned cases the experts were also asked to respond to a set of global evaluation questions. These questions are meant to capture the expert's higher-level perceptions of the system e.g., likelihood of using such summaries in routine clinical practice, overall confidence and trust in the summarization tool, and perceived risks and benefits. All responses were logged in a Google Form so they could be reviewed later. For summaries that were marked as problematic, the engineering and AI teams reviewed the issue and made relevant changes to the backend. 

Due to study design constraints, human and LLM-based evaluations employed partially distinct scoring rubrics. The rubric that was used for human experts prioritized clinical safety, interpretability, and patient relevance, whereas LLM-based evaluation focused on structured criteria like accuracy, completeness, clarity, clinical relevance, faithfulness, bias-free, and overall quality. Given the differences between the two, direct quantitative comparison between identical metrics was therefore not feasible. We describe these differences and summarize the results in section \ref{sec:val}.

\begin{table*}[t]
\centering
\caption{Descriptive statistics for model evaluation metrics (scores scaled 0--1).}
\label{tab:eval_summary_stats}
\begin{tabular}{lrrrrrrrr}
\toprule
Metric & $n$ & Mean & SD & Min & Q1 & Median & Q3 & Max \\
\midrule
Faithfulness        & 103 & 0.944 & 0.135 & 0.000 & 1.000 & 1.000 & 1.000 & 1.000 \\
Completeness        &  91 & 0.860 & 0.236 & 0.000 & 0.750 & 1.000 & 1.000 & 1.000 \\
Readability         & 103 & 0.789 & 0.125 & 0.500 & 0.750 & 0.750 & 0.750 & 1.000 \\
Helpfulness         &  95 & 0.816 & 0.089 & 0.333 & 0.833 & 0.833 & 0.833 & 1.000 \\
ProfessionalStyle   &  94 & 0.979 & 0.070 & 0.750 & 1.000 & 1.000 & 1.000 & 1.000 \\
Toxicity            & 103 & 0.000 & 0.000 & 0.000 & 0.000 & 0.000 & 0.000 & 0.000 \\
FactScore           &  93 & 0.992 & 0.044 & 0.750 & 1.000 & 1.000 & 1.000 & 1.000 \\
\bottomrule
\end{tabular}
\label{tab:judge_scores}
\end{table*}

\section{Validation Results}
\label{sec:val}
\subsection{Expert Evaluation}
\subsubsection{Evaluation Scores Clusters} We first present the analysis of responses from human experts. The responses consisted of contributions from nineteen different participants. A visual summary of the responses is given in Figure \ref{fig:hier}. It shows the hierarchical clustering of evaluation scores i.e., responses with similar characteristics are grouped together. It reveals a dominant cluster with uniformly high performance across all metrics. This corresponds to evaluations with robust and clinically reliable outputs from the summaries. A second cluster exhibits slightly lower values but the overall evaluation is still high. A third, intermediate cluster, visualized at the top, shows moderate performance across metrics, which reflects partially correct but less consistent outputs. Notably, clarity, faithfulness and accuracy displayed greater variability compared to other dimensions. In general while bias-free ratings remained consistently high across almost all responses.

\subsubsection{Analysis of Likert Scale Responses}
The Likert diverging stacked bar chart in Figure \ref{fig:likert_eval} shows an alternative view of the responses. Following standard practice for interpreting Likert scale \cite{sullivan2013analyzing}, 4-5 responses are treated as positive, 3 as neutral, and 1-2 as negative. The figure shows that clinical users overwhelmingly rated the AI-generated summaries positively across all criteria, with the vast majority of responses falling in the 4–5 positive range. Neutral ratings appear in modest amounts across criteria, while negative ratings (1–2) are nonexistent. Accuracy, Completeness, Clinical Relevance, Faithfulness, and Bias-Free language all exhibit over 75–85$\%$ positive responses. This implies consistent user satisfaction with both the clinical and linguistic quality of the summaries.

\begin{figure}[t]
    \centering
    \includegraphics[width=\linewidth]{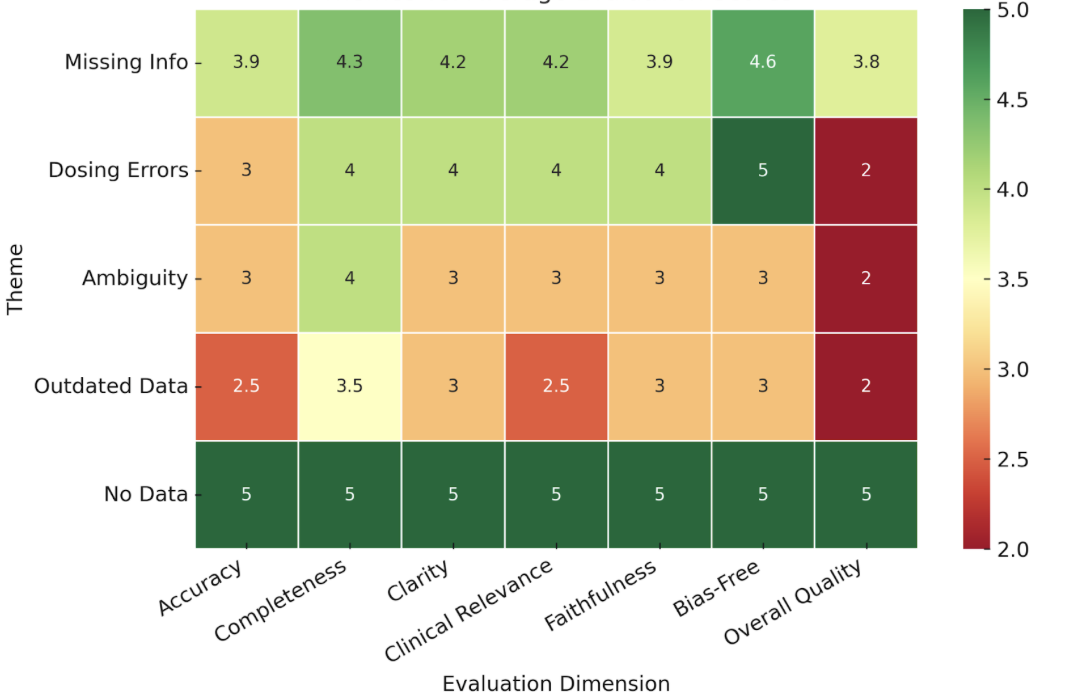}
    \caption{Theme-level performance across evaluation dimensions}
    \label{fig:themes}
\end{figure}

\subsubsection{Thematic Clustering}
To better understand the types of errors and omissions perceived by clinical evaluators, we conducted thematic clustering of written feedback on the AI-generated summaries. Thematic clustering is an unsupervised learning approach that partitions data into groups by optimizing similarity in latent semantic representations, thereby capturing shared underlying themes rather than surface-level features \cite{braun2006using}. The heatmap in Figure \ref{fig:themes} highlights a clear separation between safe and unsafe failure modes in system performance across evaluation dimensions. “No Data” cases achieve perfect scores, indicating strong guardrails when the system appropriately abstains, while “Missing Info” is handled relatively well with only minor degradation. In contrast, “Dosing Errors” emerge as a potential failure mode, showing sharp drops in overall quality despite high bias-free scores. This suggests clinically unsafe outputs rather than biased ones. “Ambiguity” leads to moderate but consistent declines across dimensions. This reflects difficulty in resolving uncertainty. “Outdated Data” primarily affects accuracy and clinical relevance. This may imply that it produces well-structured but incorrect outputs. Overall, the results suggest that whenever there is performance degradation, it may be driven less by hallucination and more by failures in reasoning and data validity.

To further investigate why we are seeing these differences, We also did manual analysis of how different error themes in AI-generated patient summaries impact evaluation scores across multiple dimensions. Perception of missing Information leads to moderately reduced scores. This suggests that clinicians notice omissions but still find the remaining content usable. “Dosing Errors” seem produce a sharper drop in Overall Quality (score of 2) despite otherwise decent scores. Further investigation revealed that this is because the medication information that the rater was expecting was not available to the AI model. “Ambiguity” results in consistently middling scores (around 3). “Outdated Data” triggers some of the lowest ratings (2.5–3). It was discovered that this was because there is no clear indication on what is the time range of data that was used. Lastly, overall, the pattern shows that content accuracy, recency, and medication correctness are the strongest drivers of clinician trust, while missing information and ambiguity reduce usefulness without fully breaking confidence. After this round of testing was completed, the issues that resulted in the errors have been addressed and the models have been updated in preparation for a second test prior to go-to-market launch.

\subsection{Automated Evaluation}
In addition to manual review of the patient data, we also evaluated the AI generated summaries via LLM-as-a-judge where we used Mistral as the main evaluator. A total of 103 AI-generated patient summaries were given to the Mistral for evaluation. A summary of the results is given in Table \ref{tab:judge_scores}. Across the 103 evaluated summaries, performance was strongest and most consistent for FactScore (mean=0.992, median=1.0) and ProfessionalStyle (mean=0.979, median=1.0), with Toxicity uniformly 0.0 across all records, indicating no detected harmful content. Faithfulness was high overall (mean=0.944; median=1.0) but included rare failures (min=0.0). In contrast, Completeness showed the greatest variability (SD=0.236; min=0.0) despite a median of 1.0. At the surface it may seem to suggest that there is a meaningful subset of summaries that omit substantial clinical information. However, detailed manual inspection of all such summaries revealed that the low score was mainly because of the name of the patient was omitted after the first sentence. Readability and Helpfulness were moderate-to-high (means 0.789 and 0.816, respectively), with tighter dispersion for helpfulness.

\begin{figure}[t]
    \centering
    \includegraphics[width=\linewidth]{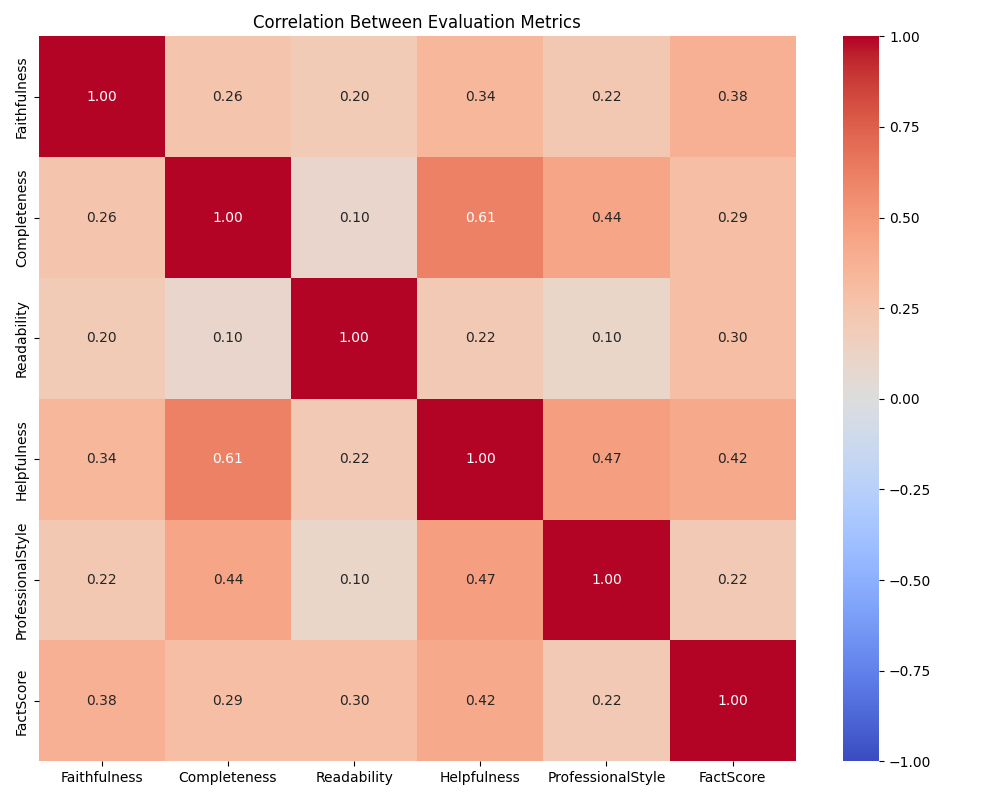}
    \caption{Correlation matrix showing relationships among evaluation metrics for LLM-generated patient summaries.}
    \label{fig:correlation_heatmap}
\end{figure}

We also performed correlation analysis of the automated evaluation responses. These are shown in Figure \ref{fig:correlation_heatmap} which indicates that the evaluation metrics capture complementary, non-redundant dimensions of quality in LLM-generated patient summaries. Helpfulness emerges as a central construct, showing the strongest association with Completeness (r = 0.61) and moderate correlations with Professional Style (r = 0.47), FactScore (r = 0.42), and Faithfulness (r = 0.34), suggesting that summaries perceived as helpful tend to balance coverage, clinical appropriateness, and factual correctness. In contrast, Readability exhibits weak correlations with other metrics (r $\leq$ 0.22), indicating that surface fluency is largely independent of substantive clinical quality. Notably, the modest correlation between Completeness and Faithfulness (r = 0.26) highlights a potential trade-off between content coverage and source adherence, underscoring the risk of over-generation in clinical summarization. Overall, these findings imply that the use of multiple metrics makes sense and one should optimze for all such metrics.

\begin{table*}[h!]
    \centering
    \small
    \renewcommand{\arraystretch}{1.5} % Adds padding to rows
    \begin{tabular}{|p{3cm}|c|p{12cm}|}
        \hline
        \textbf{Error Category} & \textbf{Score} & \textbf{Evaluator Explanation (Reason for Failure)} \\
        \hline
        Temporal Error & 0.75 & The response incorrectly states that the patient completed the assessment on Nov 21st, when the data shows it was completed on Nov 21st, \textbf{2025}. (Model confused future vs. past). \\
        \hline
        Factual Hallucination & 0.75 & The summary claims the patient's weight was recorded at 93kg, which is not accurate according to the provided data where the weight was recorded but the value was different. \\
        \hline
        Status Confusion & 0.75 & Mentions an \textit{upcoming} chemotherapy appointment, which is incorrect as the patient has \textbf{already completed} this appointment according to the logs. \\
        \hline
    \end{tabular}
    \caption{\textbf{Analysis of Faithfulness Failures.} Representative examples of the three primary error modes: temporal confusion (future dates), hallucination of specific values, and misinterpreting task status.}
    \label{tab:faithfulness_errors}
\end{table*}

We also performed a manual evaluation of all 23 instances where the Faithfulness Score was less than 1.0. Our hypothesis was that this would mainly include cases where the model may have hallucinated or made a factual error. The analysis revealed that most errors were either subtle factual inaccuracies or unsupported inferences, rather than wholesale hallucinations. Three main categories of errors were found: (i) Temporal Errors reflected summaries that were largely accurate but contained incorrect temporal details (e.g., wrong dates or sequencing of events). (ii) Factual Hallucinations e.g., potentially misreported numerical values or assertions not explicitly supported by the source data, such as inferred medication non-adherence, symptom resolution, or clinical trends. There were also several failures involved over-interpretation, where the model characterized symptoms as improving, declining, resolved, or clinically significant without sufficient evidence. 

A smaller subset of errors stemmed from instructional non-adherence i.e., omitting required patient identifiers, or violating formatting constraints. Upon closer inspection it was discovered that the prompt has asked to explicitly identify the patient by name in each sentence. Overall, these findings indicate that faithfulness errors in LLM-generated patient summaries are most commonly driven not by factual errors but by non-adherence to certain instructions. The main factual problems were related to temporal consistency checks. The various types of error are summarized in Table \ref{tab:faithfulness_errors} with examples of each error. There were a few examples which were defined by concurrent failures in faithfulness ( $<$ 0.9), completeness ( $<$ 0.8), and factual accuracy (FactScore $<$ 0.9). Analysis of these examples revealed that the  underperformance was not observed as none of the examples scored below 0.7 overall. 

\begin{figure}[t]
    \centering
    \includegraphics[width=\linewidth]{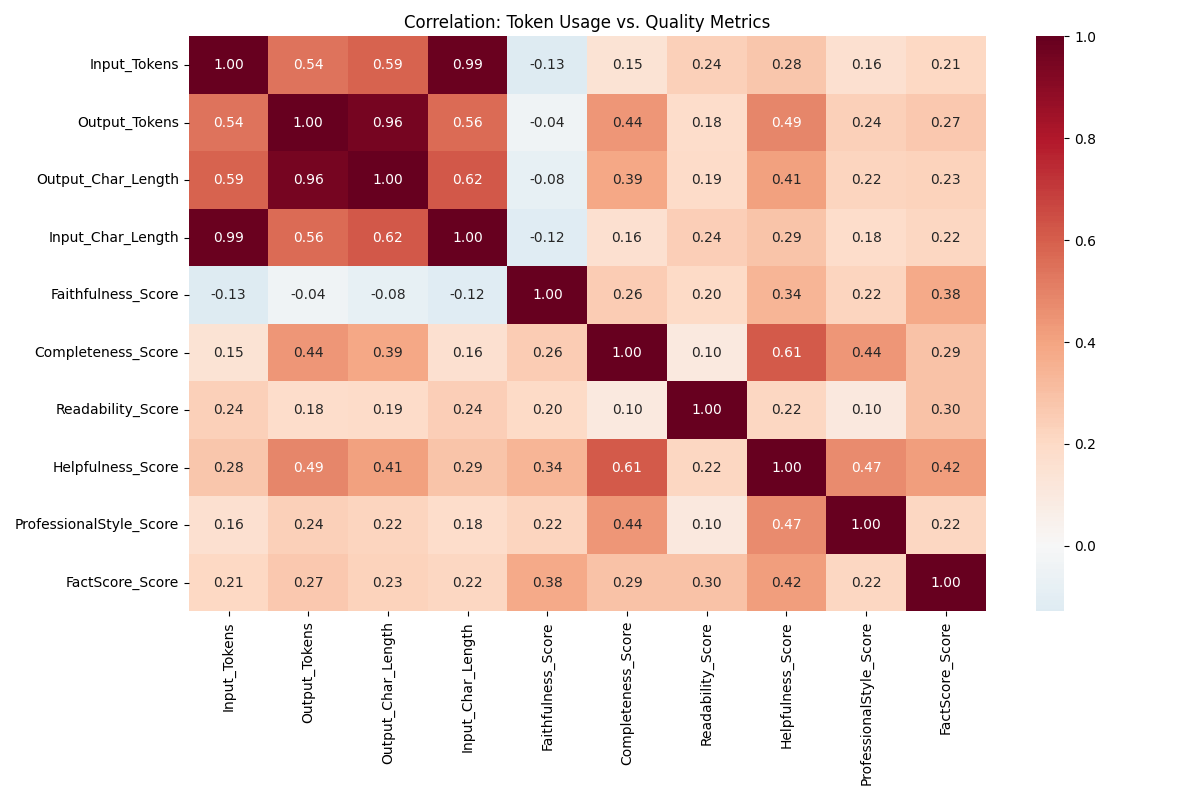}
    \caption{Correlation matrix illustrating relationships between input and output token usage and evaluation metrics for LLM-generated patient summaries.}
    \label{fig:token_correlation_matrix}
\end{figure}

We also explored if there was any relationship between the length of the summary and the score given by the LLM. This was done via correlation analysis between the token length and the metric of interest. This analysis revealed that increased input complexity and longer generated outputs are moderately associated with higher perceived utility but not with improved factual correctness. Output length (tokens and character count) showed moderate positive correlations with Completeness (r $\approx$ 0.39–0.44) and Helpfulness (r $\approx$ 0.41-0.49). This is not surprising since longer summaries tend to cover more information and thus would be judged as more useful. In contrast, Faithfulness exhibited weak negative correlations with input and output length (r $\approx$ -0.08 to -0.13). This indicates that longer or more complex summaries do not necessarily improve quality. FactScore and Professional Style show only weak positive associations with token usage (r $\approx$ 0.21-0.27). Readability is however largely independent of length. Overall, these results highlight a trade-off in which increased verbosity improves perceived completeness and helpfulness but does not necessarily confer gains in factual accuracy.

\begin{figure*}[t]
    \centering
    \includegraphics[width=\linewidth]{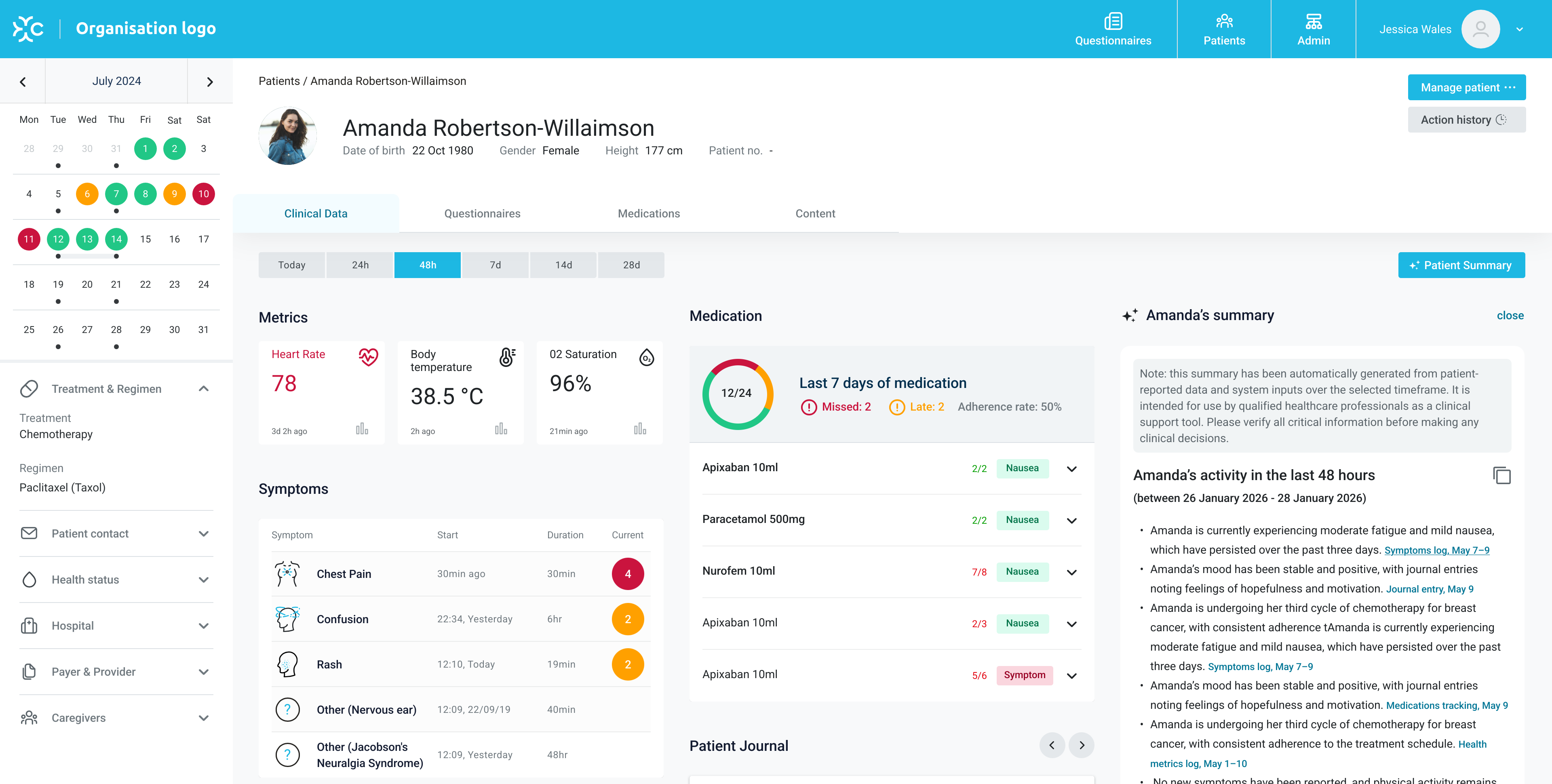}
    \caption{AI Generated Summary with Source Data}
    \label{fig:aisummary}
\end{figure*}

\section{Summarization Improvement}
\label{sec:improve}
Insights from the overall evaluation results, along with a detailed review of individual summaries that fell below the performance threshold, were used to iteratively improve summary quality. Feedback from both human experts and automated evaluation metrics informed refinements and updates to the prompt that is used to generate the summary. The goal is to enable more accurate, complete, and clinically relevant outputs. Additionally, transparency and traceability were enhanced by ensuring that each generated summary is now accompanied by the underlying source data used in its creation. This allows for easier validation, supports error identification, and strengthens trust in the system by making the connection between input data and generated output explicit.

Figure \ref{fig:aisummary} shows the healthcare personnel facing part of the platform after updates based on the suggestions. It integrates longitudinal clinical data, symptoms, medication adherence, and patient-reported inputs into a unified dashboard, complemented by an AI-generated summary for rapid situational awareness. It visualizes real-time and recent metrics (e.g., vitals, symptoms, adherence trends) while contextualizing them through a narrative summary grounded in the underlying source data, enabling clinicians to quickly assess patient status, track treatment progression, and identify potential risks. By combining structured data, patient activity logs, and explainable AI outputs, the system supports informed clinical decision-making while maintaining transparency and traceability of the information used to generate insights. It should be noted that all patient information displayed in this interface is synthetic or de-identified and does not represent real patient data.

\section{Current \& Future Work}
The results of evaluation demonstrated that the LLM generates patient summaries with high factual accuracy, appropriate clinical tone, and minimal safety risk, while exhibiting more variability in completeness and perceived usefulness. Correlation analyses show that helpfulness acts as an integrative quality dimension, aligning with completeness, professional style, and factual correctness, whereas readability remains largely independent of substantive clinical quality. Increased input complexity and longer outputs are associated with improved completeness and helpfulness but do not yield gains in faithfulness, highlighting a verbosity–grounding trade-off. Error analyses indicate that most failures arise from subtle distortions or over-interpretations rather than overt hallucinations, with rare but severe cases reflecting compounding breakdowns in coverage and factual adherence. Collectively, these findings support the use of multi-metric evaluation and conservative, grounding-focused optimization strategies for deploying LLM-generated patient summaries in safety-critical clinical settings. The updates to the systems described in section \ref{sec:improve} have been internally tested and are scheduled to be tested by a group of human domain experts.

\bibliographystyle{IEEEtran}
\bibliography{references}

\end{document}